\documentclass[10pt]{article}

\usepackage[pdftex]{graphicx}
\graphicspath{{./pdf/}}

\usepackage{csagh}
\usepackage{caption}
\usepackage{subcaption}

\begin{document}
\begin{opening}

\title{Adaptation of domain-specific transformer models with text oversampling for sentiment analysis of social media posts on Covid-19 vaccines}

\author[Delhi Technological University,
           e-mail: \URL{anmolbansal_2k18it025@dtu.ac.in}]
       {Anmol Bansal}
\author[Delhi Technological University,
           e-mail: \URL{arjunchoudhry_2k18it031@dtu.ac.in}]
       {Arjun Choudhry}
\author[Delhi Technological University,
           e-mail: \URL{anubhavsharma_2k18it029@dtu.ac.in}]
       {Anubhav Sharma}
\author[Delhi Technological University,
           e-mail: \URL{seba_406@yahoo.in}]
       {Seba Susan}

\begin{abstract}
  As the name suggests... an English abstract.
\end{abstract}

\begin{abstract}
Covid-19 has spread across the world and several vaccines have been developed to counter its surge. To identify the correct sentiments associated with the vaccines from social media posts, we fine-tune various state-of-the-art pre-trained transformer models on tweets associated with Covid-19 vaccines. Specifically, we use the recently introduced state-of-the-art pre-trained transformer models RoBERTa, XLNet and BERT, and the domain-specific transformer models CT-BERT and BERTweet that are pre-trained on Covid-19 tweets. We further explore the option of text augmentation by oversampling using Language Model based Oversampling Technique (LMOTE) to improve the accuracies of these models, specifically, for small sample datasets where there is an imbalanced class distribution among the positive, negative and neutral sentiment classes. Our results summarize our findings on the suitability of text oversampling for imbalanced small sample datasets that are used to fine-tune state-of-the-art pre-trained transformer models, and the utility of domain-specific transformer models for the classification task.
\end{abstract}


\end{opening}
\textbf{Note: The paper has been accepted for publication in Computer Science journal: http://journals.agh.edu.pl/csci}

\section{Introduction}

Covid-19 vaccines were developed in response to the widespread devastation caused by the Covid-19 pandemic which started in December 2019 and has been on the rise ever since. These vaccines have elicited a mixed response among the general public, and these reviews help us to understand just how these vaccines have affected people emotionally. Social media is the primary platform for finding solutions to health-based queries related to the Covid-19 pandemic [2]. One of the best ways to determine public opinion is to survey Twitter which has millions of users every day and receives lots of tweets on vaccines from all over the world [35]. Most of the researchers who have used machine learning to analyze Covid-19 vaccine-related tweets have adopted unsupervised techniques to determine the sentiments of tweets [21]. A majority of researchers use the rule-based Valence Aware Dictionary for sEntiment Reasoning (VADER) [15] which assigns a sentiment score to each tweet in an unsupervised manner [21, 23, 39]. Other unsupervised techniques used for Twitter sentiment analysis include AFINN [28] and TextBlob [27]. However, the unsupervised techniques are dependent on pre-defined rules meant for general sentiment analysis that may not work out for text data related to the ongoing pandemic. Supervised learning methodologies overcome the drawback of unsupervised techniques by learning appropriate text patterns that distinguish between positive, negative and neutral sentiments [45, 32, 43]. 
Supervised learning for text classification is achieved in the present times using transformers that are now popularly replacing Long Short-Term Memory (LSTM) [26] and Convolutional Neural Network (CNN) [11] in various Natural Language Processing (NLP) tasks [13, 33]. Transformer models can be used for sequence modeling to predict the next word in a sentence, and are usually trained on a large corpus such as Wikipedia or Brown Corpus [41, 9]. These pre-trained models are generalized, and are usually fine-tuned for downstream tasks such as classification and text generation.\par
One of the challenges faced in supervised machine learning is the small size of the dataset which leads to less accurate models. Such datasets are called small sample datasets (SSD) [16]. The situation is complicated when the class distribution is imbalanced with the majority class samples outnumbering the minority class samples [35]. SSD provides fewer examples for the model to identify patterns from, and therefore results in less accurate models. An imbalanced dataset can also be detrimental to the model as it results in biased results towards the majority class and therefore results in wrong predictions. This paper aims to explore solutions to the class imbalance problem associated with the sentiment categories of Covid-19 related tweets when the size of the dataset is small. We explore the viability of text-based oversampling as a possible solution. The synthetic tweets generated are from classes having a lower population in order to balance the population of all classes while increasing the number of training samples at the same time. We explore different pre-trained transformer models for supervised learning from an imbalanced, small sample dataset containing tweets on Covid-19 vaccine related discussions, and compare their performance with that of domain-specific transformer models for the classification task. Specifically, we investigate the performance of the state-of-the-art transformer models RoBERTa, BERT and XLNet as compared to the domain-specific pre-trained transformer models CT-BERT and BERTweet for sentiment analysis of Covid-19 vaccine related tweets. CT-BERT and BERTweet are pre-trained transformer models obtained after intensive training on English tweets related to the Covid-19 pandemic. On the other hand, RoBERTa, XLNet and BERT provide input embeddings for sentences written in English, and are used for generalized natural language processing. CT-BERT and BERTweet models have the advantage of being familiar with text patterns emanating from Covid-19 related discussions such as “Covid positive” which may not be understood well by the RoBERTa, BERT and XLNet models. \par
The organization of this paper is as follows. Section 2 describes some related work on text oversampling of Covid-19 related text datasets, and also reviews the LMOTE algorithm used for text oversampling in the current work, Section 3 outlines several state-of-the-art pre-trained transformer models including domain-specific transformer models, Section 4 presents the methodology, Section 5 contains a detailed analysis of the results, and Section 6 concludes the paper.

\section{Preliminaries of text oversampling}

\subsection{Text oversampling of Covid-19 related text}

The Small Sample Size (SSS) problem refers to the availability of a small number of training samples in high dimensional datasets [25]; this leads to inadequate training rendering supervised learning a challenging task. This paper uses a small subset of annotated Covid-19 tweets to observe the effects of using a smaller text dataset for fine-tuning pre-trained transformer models that are the state of the art for implementing various NLP tasks. The situation is complicated when the class distribution is uneven and the number of majority samples is much more than the number of minority samples [35]. In literature, there are several examples of text oversampling being applied to Covid-19 related social media posts due to the apparent scarcity of text samples belonging to some of the minority classes. We discuss some of these works next. In [22], Liu et al. proved that oversampling the term-frequency inverse document frequency (TF-IDF) features using Synthetic Minority Oversampling Technique (SMOTE) [10] improved the results of Covid-19 vaccine hesitancy prediction. Support Vector Machine (SVM) was used for the classification. SMOTE was also used in [4] to oversample word embeddings for sentiment analysis of Arabic tweets related to Covid-19 conspiracy theories. A recent work [36] investigated ensemble models for the classification of Covid-19 infodemic tweets oversampled using SMOTE. Mohsen et al. [31] recommended text oversampling using SMOTe Edited Nearest Neighbor (SMOTEENN) for the sentiment analysis of Arabic tweets related to Covid-19 quarantine. Random oversampling was performed in [5] for detecting Covid-19 misinformation on Twitter. 

\subsection{A review of the LMOTE algorithm for text oversampling}

Language Model-based Oversampling Technique (LMOTE), proposed by Leekha et al. in 2020 [20] is a language modelling based synthetic datapoints generation approach for tackling the problem of class imbalance in natural language processing tasks. Previous synthetic datapoints generation approaches for tackling class imbalance, like SMOTE and its variants [42], lack the ability to allow for proper qualitative analysis of the generated synthetic data points since the synthetic samples were generated in the Euclidean space. This made it difficult to concretely judge the semantic and contextual validity of the generated synthetic data points. Unlike SMOTE and its variants, LMOTE works specifically on textual data, and the generated synthetic data points using LMOTE allow for more concrete and intuitive balancing of the dataset. \par
In our current work on Covid-19 vaccine sentiment analysis, there are three classes of sentiments: positive, negative and neutral. The neutral tweets are large in number since a lot of people tweet about generic information regarding vaccines without expressing any sentiments. Hence neutral sentiment is the majority class in our problem. The algorithm for text oversampling of the tweets belonging to the minority class (positive and negative tweets in our case) using LMOTE is given below.

\begin{figure}[!ht]
\centering
\includegraphics[scale=0.9]{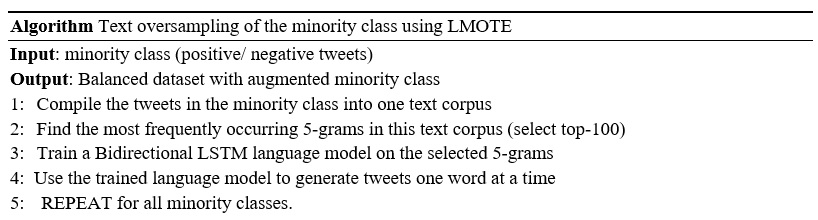}
\end{figure}

\section{Pre-trained transformer models for domain-specific tasks}

Transformers introduced by Vaswani et al. in 2017 [44] rely on the concept of self-attention that involves computation of intra-attention between positions in the input sequence. Transformer is rapidly replacing LSTM and CNN in encoder-decoder models that incorporate an attention mechanism between the encoder and decoder [7, 21]. Google’s BERT [12] and XLNet [46] are bi-directional transformer models used for learning representations for various NLP tasks, with XLNet outperforming BERT on several tasks that involve learning from long sequences [3, 46]. BERT is a transformer-based model based on masked language modeling. BERT and its advanced versions such as RoBERTa [24] and ALBERT [17] are trained on the English Wikipedia and the Brown Corpus. Pre-trained BERT models generate word embeddings that can be used for text understanding and classification. BERT can also be fine-tuned to adapt to specific tasks. XLNet is an autoregressive (AR) language model which uses permutation language modeling during the pre-training phase. Though it is similar in architecture to BERT, it differs in its pre-training objective due to which it surpasses BERT in various NLP tasks. XLNet is pre-trained using only a subset of output tokens as a target. Like BERT, pre-trained XLNet models can also be used for many other downstream tasks while also increasing the limits for sequences. In a recent work, the XLNet transformer was used successfully for sentiment analysis of unlabeled Covid-19 tweets by transfer learning [8]. The XLNet transformer was pre-trained on the US Airlines tweets dataset that was unconnected with Covid-19. 
Several language-specific models have been developed such as CamemBERT for French [29] and GottBERT for the German language [40]. These specific language models give better results than the BERT multi-language model. The generalized transformer models can be further trained on downstream tasks to create separate models for different tasks and different languages. We discuss some of these domain-specific models here. Each of them is trained on a specialized corpus, relevant to the topic at hand, and is more effective in that domain.

1. SciBERT (biomedical and computer science literature corpus): It is a BERT-based language model for performing scientific tasks. It was introduced by Beltagy et al. in 2019 [9]. 

2. FinBERT (financial services corpus): This is a pre-trained NLP model proposed by Araci in 2019 [6] for analyzing the sentiments of financial statements, and is trained using a large financial corpus. 

3. BioBERT (biomedical literature corpus): This NLP model pre-trained on biomedical corpora outperformed BERT and various state-of-the-art models in a variety of biomedical text mining tasks. It was introduced by Lee et al. in 2019 [19].

4. ClinicalBERT (clinical notes corpus): This model focuses on clinical notes and its representations using bidirectional transformers and uncovers the relationship between medical concepts and humans as discussed by Huang et al. in 2019 [14]. 

5. mBERT (corpora from multiple languages): mBERT is a single BERT model proposed by Pires et al. in 2019 that is trained on 104 different languages [37]. The languages with fewer data were oversampled and those with surplus of data were undersampled to balance the corpus.

6. patentBERT (patent corpus): The patentBERT model is a fine-tuned pre-trained BERT model for patent classification proposed by Lee et al. in 2019 [18]. The fine-tuning was done using over 2 million patents and CNN with word embeddings.

7. RoBERTa (Optimized pre-training approach for BERT): RoBERTa is a robustly optimized pre-trained model based on BERT [24]. It is implemented on PyTorch and modifies key hyperparameters of BERT, including the removal of BERT’s next-sentence pre-training objective and is trained with much larger mini-batches and learning rates. This helps it to achieve better downstream performance in the masked language modeling approach of BERT.

8. COVID-Twitter-BERT or CT-BERT (Covid-19 tweets): This is a domain-specific transformer-based model which is pre-trained on 160 million Twitter messages specifically related to Covid-19 [30]. The aim is to understand the content of social media posts related to the Covid-19 pandemic. Muller et al. proposed this model in 2020 [30] and applied it for five different classification tasks. The model gave an improvement over BERT on COVID-19 datasets but needed more pre-training to achieve similar performance on out-of-domain contents.

9.  BERTweet (English Tweets): This model is a large-scale pre-trained language model for English Tweets and has the same architecture as the base BERT model [34]. Experiments have proved that this model outperforms strong baselines RoBERTa-base and XLM-R-base on various NLP tasks. BERTweet is the first public large-scale model pre-trained on English Tweets. It is trained using the RoBERTa pre-training procedure. This model is trained using a corpus of 850 million English Tweets comprising of 845 million tweets streamed from January 2012 to August 2019 and 5 million tweets belonging to the COVID-19 pandemic.

\section{Implementation details}

In our work, we test the suitability of text data augmentation for Covid-19 vaccine-related tweets applied for fine-tuning pre-trained transformer models. Data augmentation of the minority class is a popular remedy for class imbalance [38]. Due to the significant class-imbalance among positive, negative and neutral tweets, we oversample the positive and negative tweets only (minority class), concatenating the synthetic data points to the original dataset to generate a more balanced dataset. We adapt the LMOTE model for augmenting the data that is given as input to the pre-trained transformer models. 

The dataset used is a small subset of the larger set of Covid-19 tweets presented by Gabriel Preda [1] which is annotated for positive, negative, and neutral sentiments by FullMoonDataScience. The dataset contains 6000 tweets with 3680 tweets belonging to the neutral class, 1900 tweets to the positive class, and 420 tweets to the negative class. The dataset is thus both small in size and highly imbalanced. The data format is shown in Table 1. 

\begin{table}[!ht]
\centering
\caption{Data format}
\label{tabl.1}
  \begin{tabular}{|l<{}|c|>{$}c<{$}|r|}
                                 \hline
    \textbf{column}   & \textbf{type} \\\hline
   tweetID   & integer  \\\hline
   label  & 1, 2, 3 \\\hline
   text & string \\\hline
  \end{tabular}
\end{table}

The input text sequences from Covid-19 tweets are tokenized. We use the hugging-face library in Python for the implementation of the transformer models. The text in the tweets is pre-processed by removing hashtags, links, emails, punctuations and extra spaces using regex. The sequential pre-processing steps are shown for an example tweet in Table 2. In addition to the steps shown, tabs and extra spaces are also removed.

\begin{table}[!ht]
\centering
\caption{Sequential pre-processing steps for an example tweet}
\label{tabl.1}
  \begin{tabular}{|>{\centering\arraybackslash}m{2cm}|>{\centering\arraybackslash}m{10cm}|}
                                 \hline
    \textbf{Pre-processing step}   & \textbf{Output} \\\hline
Original tweet  &  "More \#GoodNewsfrom @bopanc \&amp; @DovLieber! \#PfizerBioNTech's \#COVID \#vaccine is highly effective after just 1 dose \&amp; can be stored in ordinary freezers for up to 2 weeks, according to new data https://t.co/ ZWwi00rIU via @WSJ https://t.co/7TMIPCkkBa"\\\hline
Removing @ mentions  &  "More \#GoodNews from \&amp; ! \#PfizerBioNTech's \#COVID \#vaccine is highly effective after just 1 dose \&amp; can be stored in ordinary freezers for up to 2 weeks, according to new data https://t.co/QZWwi00rIU via https://t.co/7TMIPCkkBa"\\\hline
Removing Hash tags  &  "More from \&amp; ! 's is highly effective after just 1 dose \&amp; can be stored in ordinary freezers for up to 2 weeks, according to new data https://t.co/QZWwi00rIU via https://t.co/7TMIPCkkBa"\\\hline
Removing websites  &  "More from \&amp; ! 's is highly effective after just 1 dose \&amp; can be stored in ordinary freezers for up to 2 weeks, according to new data via "\\\hline
Removing every Punctuation Mark except - !?.  &  "More from amp ! 's is highly effective after just 1 dose amp can be stored in ordinary freezers for up to 2 weeks according to new data via "\\\hline
Removing numbers  & "More from amp ! 's is highly effective after just dose amp can be stored in ordinary freezers for up to weeks according to new data via "\\\hline

  \end{tabular}
\end{table}

A three-fold cross-validation is performed for all the models in our experimentation:- RoBERTa, XLNet and BERT, and the domain-specific CT-BERT and BERTweet. The cross-entropy loss function and Adam Optimizer are used for training the models with a learning rate of 2e-5, and five epochs. Hyperparameter settings are chosen as per the guidelines in the original papers [25]. Fig. 1 shows the process flow pipeline showing the training and testing phases.

\begin{figure}[!ht]
\centering
\includegraphics[scale=0.9]{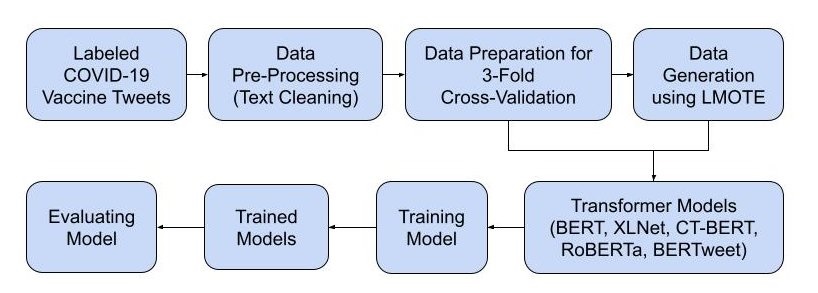}
\caption{Process flow}
\end{figure}

The tokenized text is subject to text oversampling, where the (minority classes) positive and negative tweets are oversampled using LMOTE such that the population of all three classes positive, negative and neutral are balanced. Some of the generated tweets are shown in Table 3 for reference. Neutral tweets were not oversampled since they constitute the majority class. Negative tweets were highly augmented using synthetic samples to match the population of the neutral class. 

\begin{table}[!ht]
\centering
\caption{Some instances of tweets generated by text oversampling using LMOTE}
\label{tabl.1}
  \begin{tabular}{|>{\centering\arraybackslash}m{7cm}|>{\centering\arraybackslash}m{1.75cm}|}
                                 \hline
    \textbf{text generated}   & \textbf{label} \\\hline
reporting cases of new cases in toronto closed business church no family no hope canada get out of syria effective and tech   & positive  \\\hline
the commission has secured million additional doses of vaccine bringing the total number of doses secured to billion europeans will have had the  & positive \\\hline
home no family no business no church lent no hope until jab unproven vaccine order russia get out of syria that we have no idea of the & negative \\\hline
company trial participants have died had received amp s s now not not worry to show for any but i know that experienced after the first dose of my & negative \\\hline
  \end{tabular}
\end{table}

The augmented and balanced dataset is applied for fine-tuning of the pre-trained models RoBERTa\footnote{https://github.com/facebookresearch/fairseq/blob/main/examples/roberta/README.md}, BERT\footnote{https://github.com/google-research/bert} and XLNet\footnote{https://github.com/zihangdai/xlnet}, and the domain-specific pre-trained models BERTweet\footnote{https://github.com/VinAIResearch/BERTweet} and CT-BERT\footnote{https://github.com/digitalepidemiologylab/covid-twitter-bert}.

\section{Results}

Our experiments were performed in Python version 3.8 on a 2.8 GHz Intel core PC. We have made our source code available online\footnote{https://github.com/Ace117MC/transformer-models-covid} for research purposes. The Covid-19 tweets were pre-processed and tokenized as per the procedure outlined in Section 4. In order to explore the effects of text oversampling on our small sample dataset, we performed the experiment twice, one with text oversampling and one without. The text data was used to fine-tune the pre-trained models RoBERTa, BERT and XLNet, and the domain-specific pre-trained models CT-BERT and BERTweet. The transformer models were trained using three-fold cross-validation on three different 80-20 splits of the dataset, and then we obtained the mean of the performance metrics over the three runs. 

\subsection{Results without text oversampling}

The performance metrics - test accuracy, F1-score and Mathew's correlation coefficient (MCC) are summarized in Table 4 for the five transformer models - RoBERTa, BERT, XLNet, CT-BERT and BERTweet in the absence of text oversampling. The corresponding receiver operating characteristic (ROC) curves (with Area Under Curve (AUC) readings) are plotted for the three sentiment classes in Fig. 2 (a, b, c).

\begin{figure}[!ht]
\centering
\begin{subfigure}[b]{0.3\textwidth}
\centering
\includegraphics[scale=0.2]{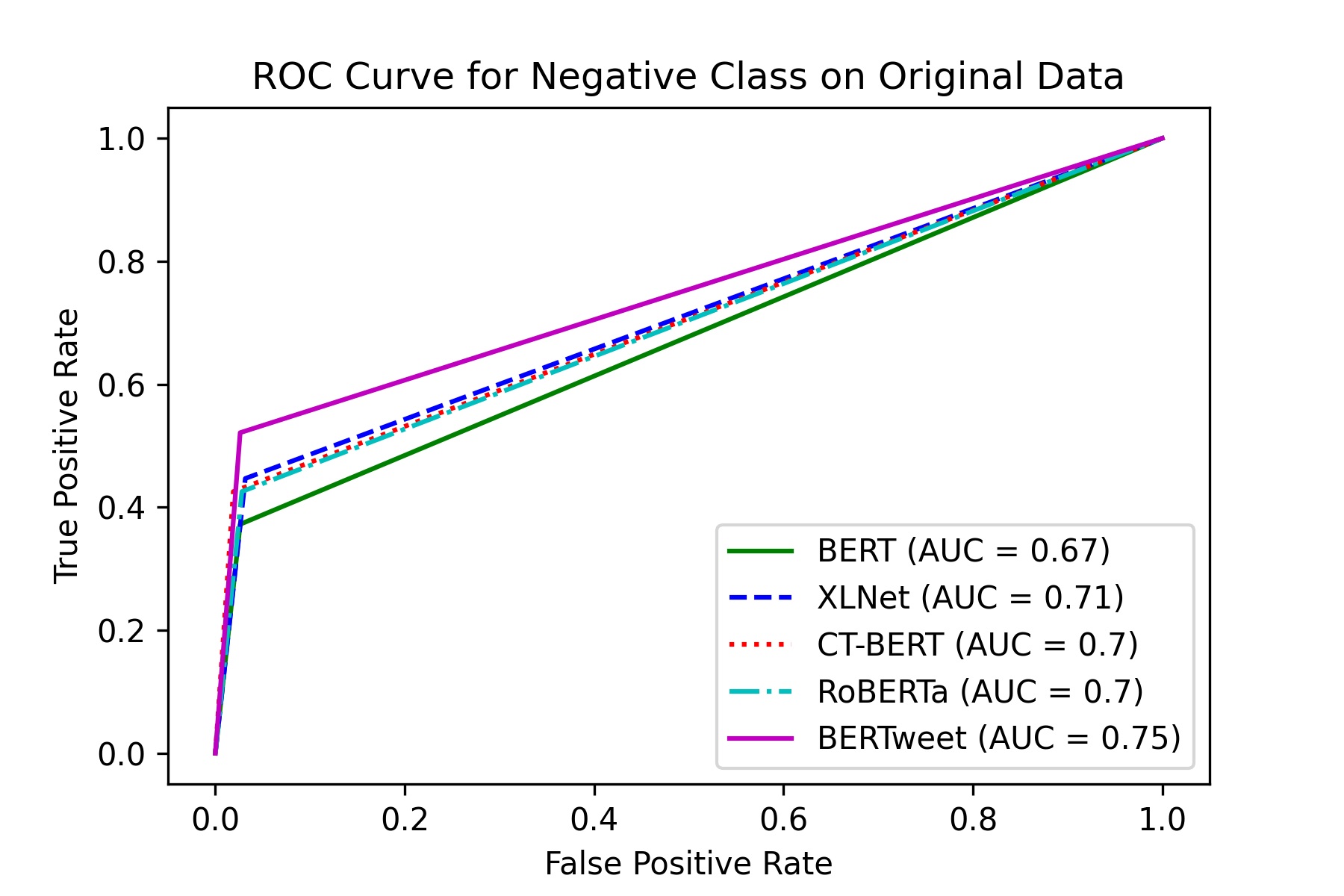}
\caption{Negative class}
\end{subfigure}

\begin{subfigure}[b]{0.3\textwidth}
\centering
\includegraphics[scale=0.2]{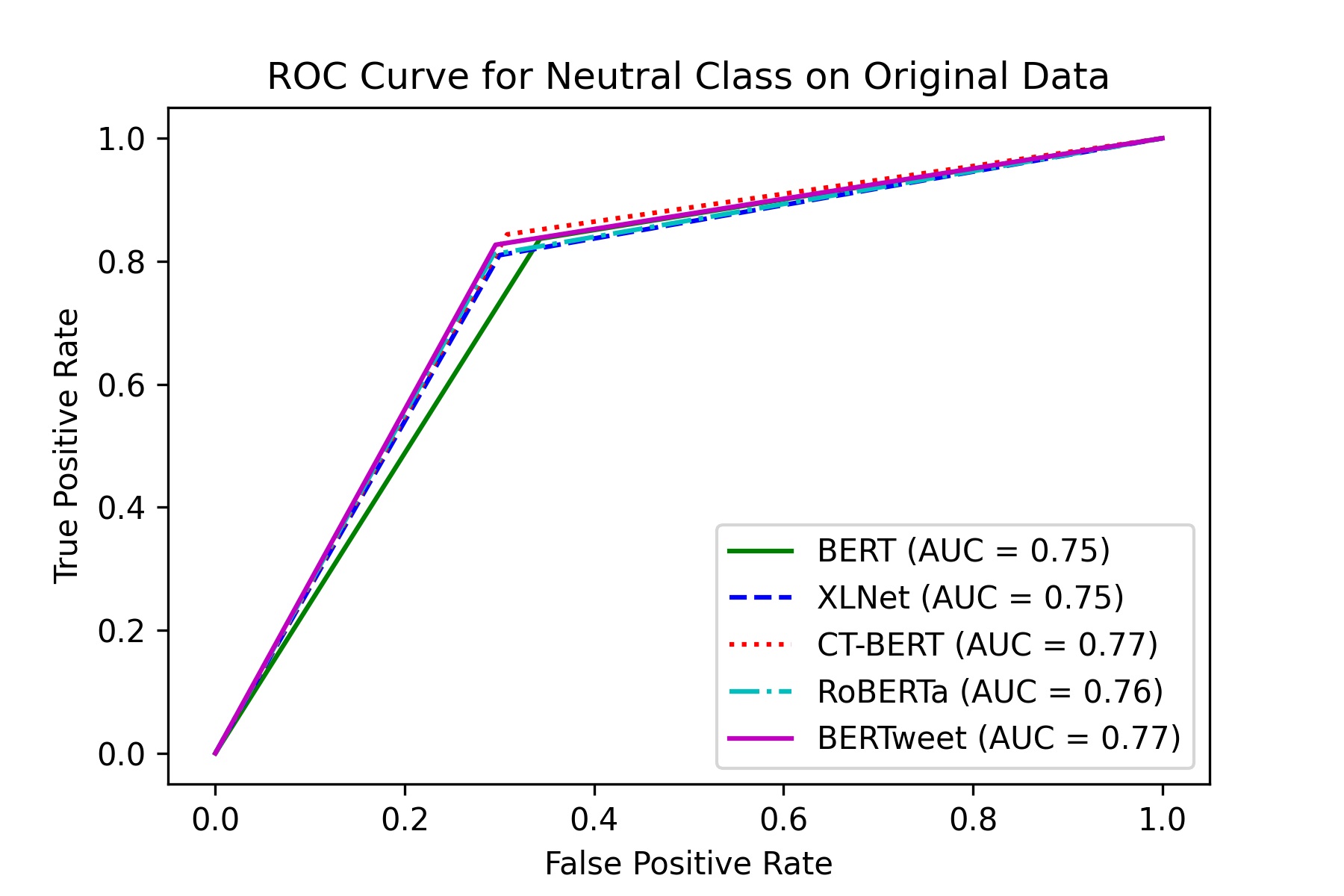}
\caption{Neutral class}
\end{subfigure}

\begin{subfigure}[b]{0.3\textwidth}
\centering
\includegraphics[scale=0.2]{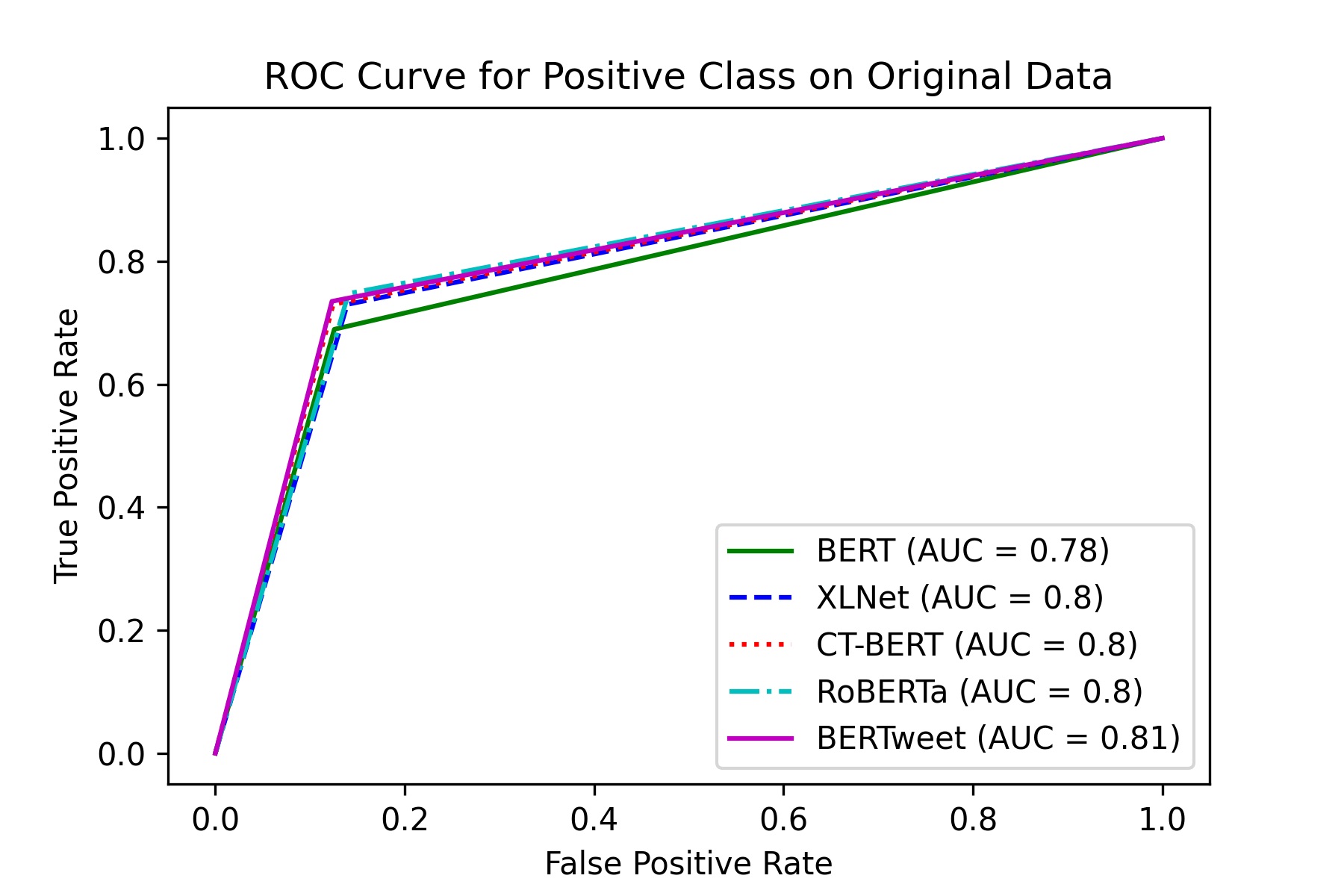}
\caption{Positive class}
\end{subfigure}
\caption{ROC curves and AUC values for different models (w/o text oversampling)}
\end{figure}

\begin{table}[!ht]
\centering
\caption{Test accuracy, F1-Score and Mathew's correlation coefficient (MCC) for the five transformer models (w/o text oversampling)}
\label{tabl.1}
  \begin{tabular}{|l<{}|c|>{$}c<{$}|r|}
                                 \hline
    \textbf{Model}   & \textbf{Accuracy}  & \textbf{F1-score}  & \textbf{MCC} \\\hline
   BERT   & 75.06\%  & 0.63 & 0.51 \\\hline
   XLNet   & 75.64\%  & 0.67 & 0.53 \\\hline
  RoBERTa   & 76.44\%  & 0.67 & 0.55 \\\hline
 CT-BERT   & 77.70\%  & 0.70 & 0.57 \\\hline
 BERTweet   & 77.25\%  & 0.70 & 0.56\\\hline
  \end{tabular}
\end{table}

\begin{table}[!ht]
\centering
\caption{Class-wise performance analysis of the transformer models for positive, negative and neutral sentiments (w/o text oversampling)}
\label{tabl.1}
  \begin{tabular}{|l<{}|c|>{$}c<{$}|c| |c| |c|}
                                 \hline
    \textbf{Model} & \textbf{Category} & \textbf{Precision} & \textbf{Recall} & \textbf{F1-score} & \textbf{MCC}  \\\hline
   BERT & neutral  & 0.79  & 0.83 & 0.81 & 0.49 \\\hline
   BERT & positive  & 0.72  & 0.69 & 0.70 & 0.57 \\\hline
   BERT & negative  & 0.44  & 0.34 & 0.38 & 0.34 \\\hline
   XLNet & neutral  & 0.82  & 0.80 & 0.81 & 0.51 \\\hline
   XLNet & positive  & 0.70  & 0.74 & 0.72 & 0.58 \\\hline
   XLNet & negative  & 0.50  & 0.48 & 0.49 & 0.45 \\\hline
   RoBERTa & neutral  & 0.82  & 0.81 & 0.81 & 0.53 \\\hline
   RoBERTa & positive & 0.71  & 0.75 & 0.73 & 0.6 \\\hline  
   RoBERTa & negative  & 0.51  & 0.45 & 0.48 & 0.44 \\\hline  
    CT-BERT & neutral  & 0.81  & 0.84 & 0.82 & 0.55 \\\hline
    CT-BERT & positive  & 0.75  & 0.70 & 0.72 & 0.61 \\\hline
    CT-BERT & negative  & 0.58  & 0.54 & 0.56 & 0.53 \\\hline
 BERTweet & neutral  & 0.81  & 0.82 & 0.81 & 0.53 \\\hline
   BERTweet & positive  & 0.73  & 0.73 & 0.73 & 0.6 \\\hline
   BERTweet & negative  & 0.56  & 0.52 & 0.54 & 0.51 \\\hline
  \end{tabular}
\end{table}

It is observed from Table 4, that the domain-specific CT-BERT and BERTweet models show a better performance as compared to the RoBERTa, BERT and XLNet models in terms of test accuracy, F1-score and MCC. RoBERTa is observed to be better than BERT and XLNet, while BERTweet proved to be almost as good as CT-BERT and better than RoBERTa. The reason for the better performance of CT-BERT and BERTweet models is that they are specifically trained on COVID-19 tweets, and are hence familiar with text patterns related to the pandemic-related discussions. 

The detailed class-wise precision, recall, F1-score and MCC readings are presented in Table 5 for the five transformer models. There are three sentiment classes- neutral which is the majority class, and positive and negative which are minority classes, with the negative tweets being very small in number. As expected, the results are biased, with the neutral and positive classes performing better than the negative class, as observed from both Table 5 (F1-score, MCC) and Fig. 2 (AUC). The negative tweets being very low in number are highly mis-classified as evident from the poor performance of the negative class. CT-BERT is the best performer out of all the five models, followed by BERTweet, in terms of test accuracy, F1-score and MCC, followed by RoBERTa and XLNet. The accuracies of CT-BERT and BERTweet for the minority class (negative sentiment) are found significantly higher than the other models in Table 5, verifying that domain-specific transformer models mitigate the effect of the class imbalance to a certain extent, even in the absence of text augmentation.

\subsection{Results with text oversampling}
 
We next demonstrate the effects of text oversampling (of the positive and negative sentiment classes) using LMOTE to investigate the suitability of text augmentation prior to the training phase. The test accuracy, F1-score and MCC values are summarized in Table 6 for the five transformer models - RoBERTa, BERT, XLNet, CT-BERT and BERTweet when text oversampling is performed using LMOTE. 

\begin{table}[!ht]
\centering
\caption{Test accuracy, F1-Score and Mathew's correlation coefficient (MCC) for the five transformer models on dataset augmented using LMOTE}
\label{tabl.1}
  \begin{tabular}{|l<{}|c|>{$}c<{$}|r|}
                                 \hline
    \textbf{Model}   & \textbf{Accuracy}  & \textbf{F1-score} & \textbf{MCC} \\\hline
   BERT   & 74.25\%  & 0.62 & 0.49 \\\hline
   XLNet   & 75.80\%  & 0.65 & 0.53  \\\hline
  RoBERTa   & 76.69\%  & 0.67 & 0.55 \\\hline
 CT-BERT   & 76.14\%  & 0.67 & 0.52\\\hline
 BERTweet   & 77.78\%  & 0.68 & 0.56  \\\hline
  \end{tabular}
\end{table}

A scrutiny of the results in Table 6 reveal a slight dip in the performance scores after text oversampling as compared to the readings in Table 4 (w/o text oversampling). This indicates that text oversampling of the minority classes for a small sample dataset will not improve the classification accuracy. For the augmented dataset, BERTweet performed best followed by RoBERTa and CT-BERT. \par
We also compare the performance of LMOTE with SMOTE; the performance scores for the five transformer models when performing text oversampling using SMOTE are compiled in Table 7. On comparing the scores of SMOTE in Table 7 with the results of LMOTE in Table 6, we note that the performance of LMOTE is found to be significantly higher than that of SMOTE. This proves that text generation by language modeling is a better option for augmentation of text corpora than the resampling strategies prevalent in data mining.

\begin{table}[!ht]
\centering
\caption{Test accuracy, F1-Score and Mathew's correlation coefficient (MCC) for the five transformer models on dataset augmented using SMOTE}
\label{tabl.1}
  \begin{tabular}{|l<{}|c|>{$}c<{$}|r|}
                                 \hline
    \textbf{Model}   & \textbf{Accuracy}  & \textbf{F1-score} & \textbf{MCC} \\\hline
   BERT   & 61.43\%  & 0.54 & 0.37 \\\hline
   XLNet   & 56\%  & 0.51 & 0.36  \\\hline
  RoBERTa   & 57.97\%  & 0.54 & 0.38 \\\hline
 CT-BERT   & 69.58\%  & 0.63 & 0.47\\\hline
 BERTweet   & 69.02\%  & 0.6 & 0.45  \\\hline
  \end{tabular}
\end{table}

\begin{figure}[!ht]
\centering
\begin{subfigure}[b]{0.3\textwidth}
\centering
\includegraphics[scale=0.2]{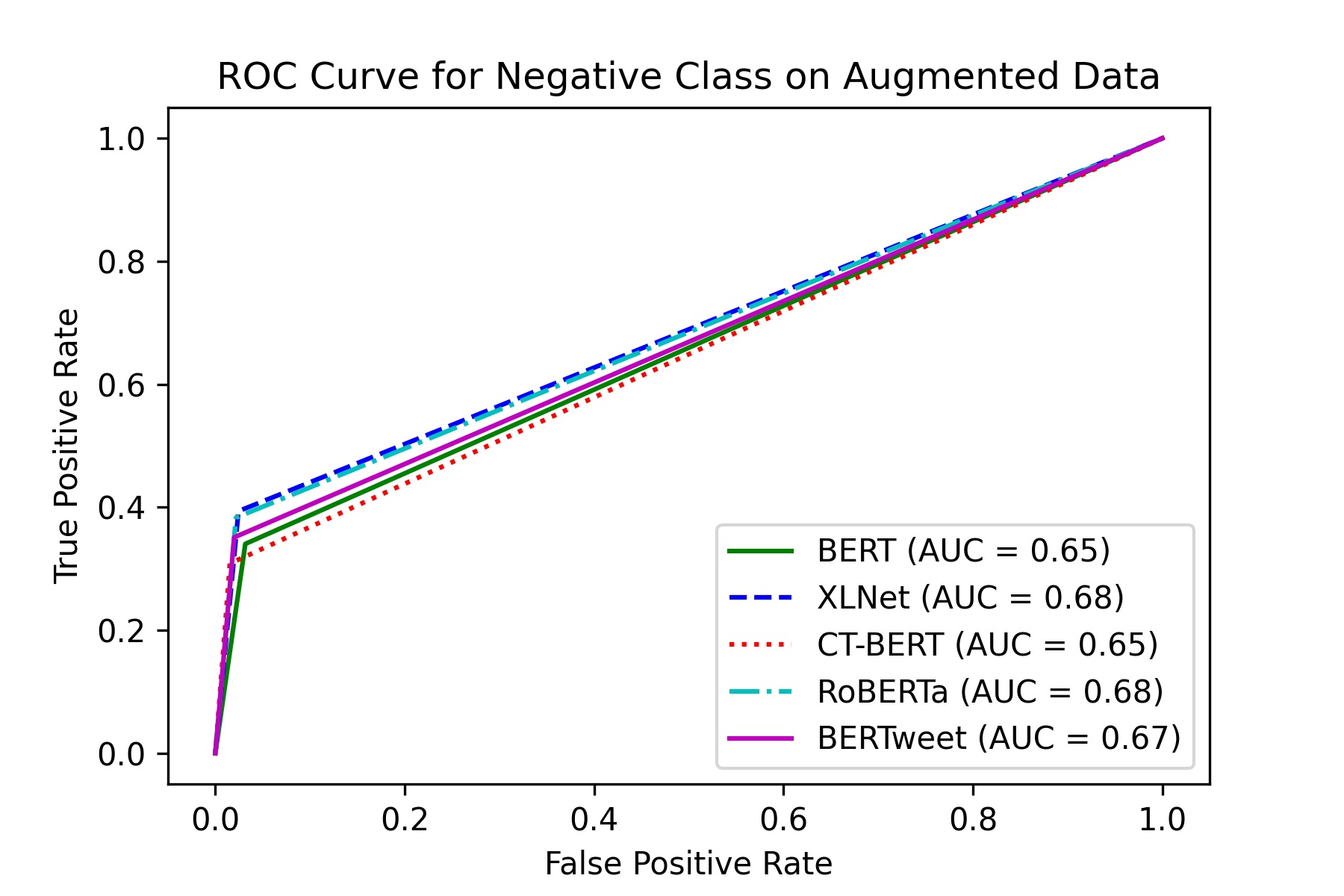}
\caption{Negative class}
\end{subfigure}

\begin{subfigure}[b]{0.3\textwidth}
\centering
\includegraphics[scale=0.2]{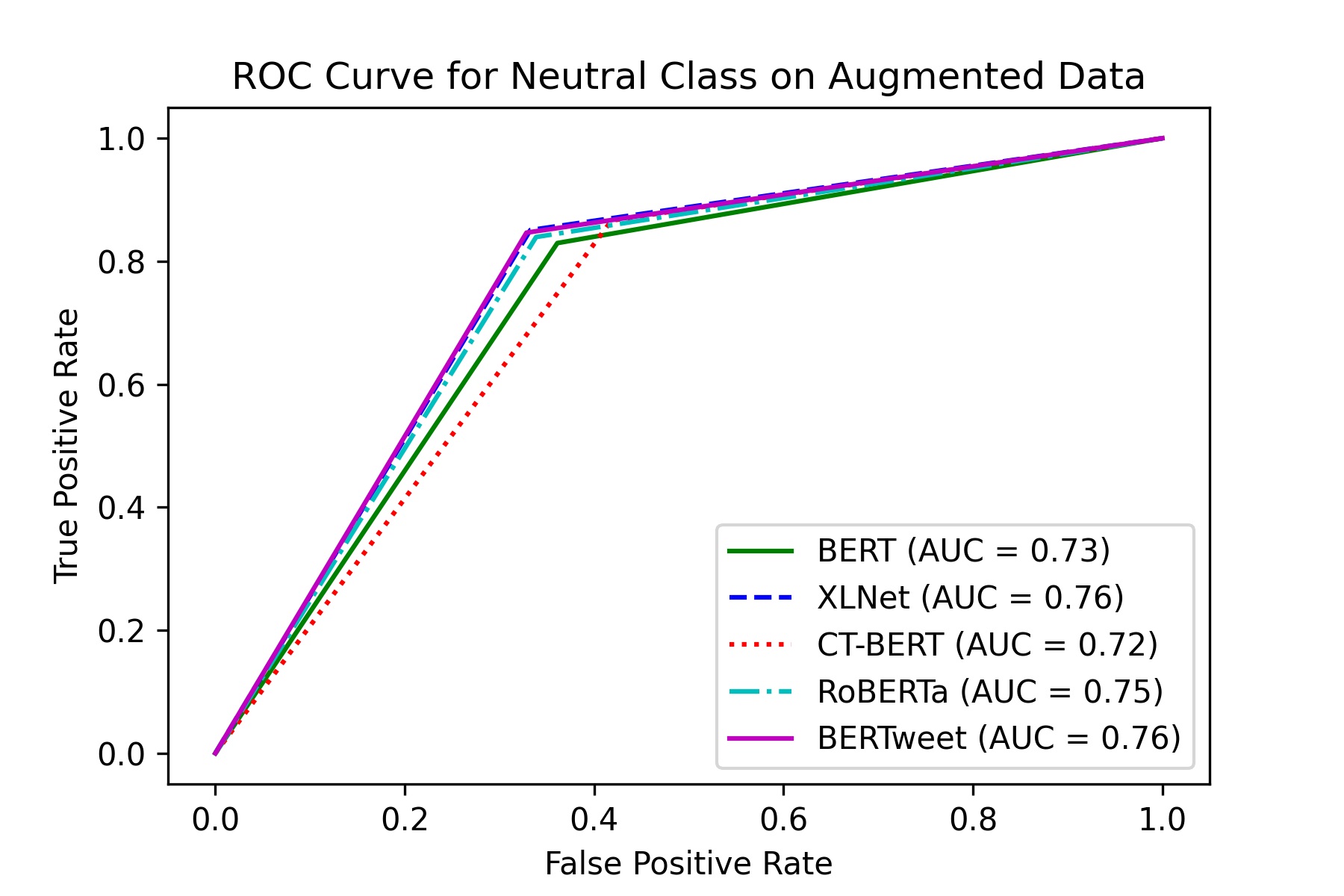}
\caption{Neutral class}
\end{subfigure}

\begin{subfigure}[b]{0.3\textwidth}
\centering
\includegraphics[scale=0.2]{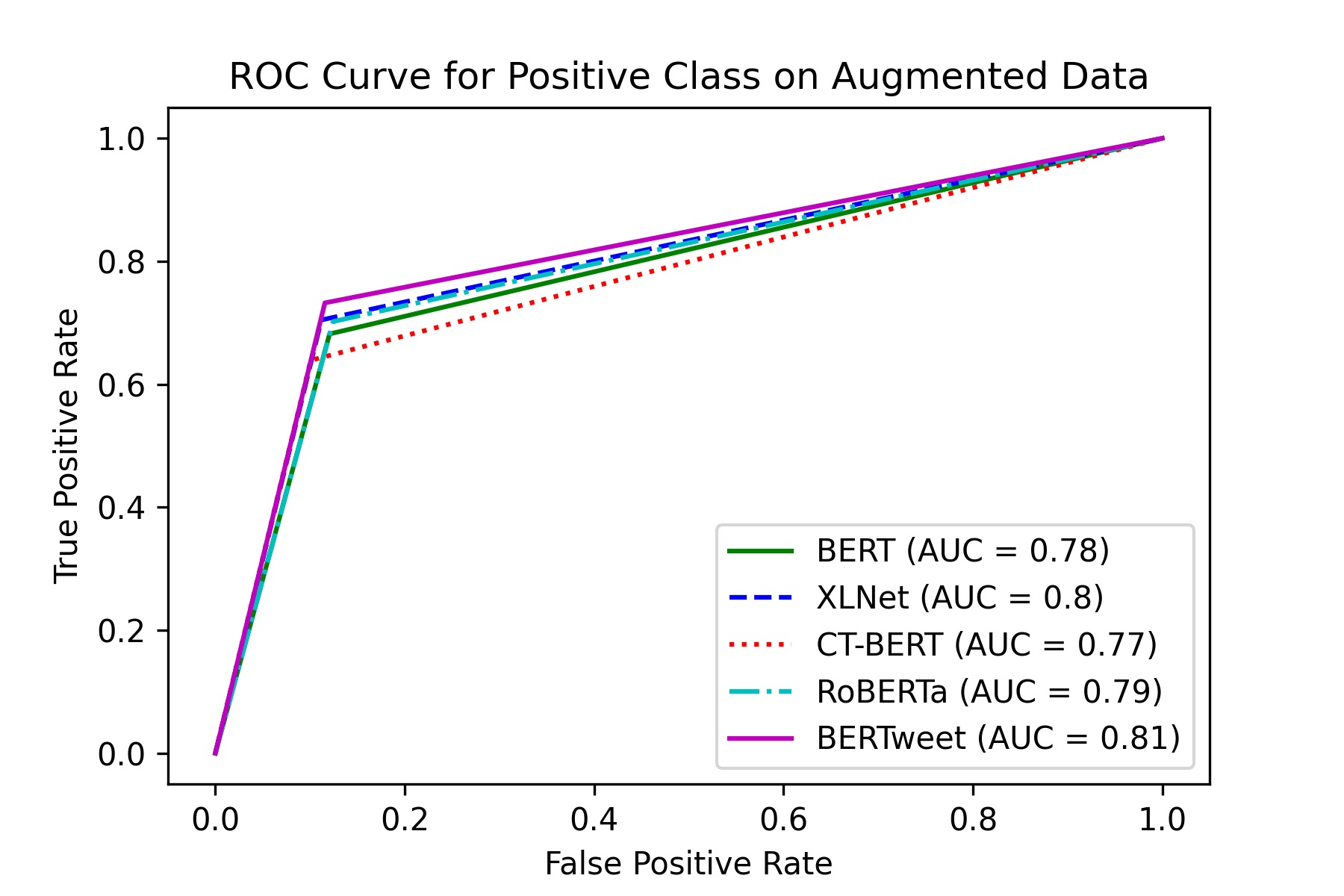}
\caption{Positive class}
\end{subfigure}
\caption{ROC curves and AUC values for different models (with text oversampling)}
\end{figure}

The detailed class-wise accuracies are presented in Table 8 for all the five transformer models, in case of text augmentation using LMOTE. The corresponding ROC curves (with AUC readings) are plotted for the three sentiment classes in Fig. 3 (a, b, c). 

\begin{table}[!ht]
\centering
\caption{Class-wise performance analysis of the transformer models for positive, negative and neutral sentiments (for dataset augmented using LMOTE)}
\label{tabl.1}
  \begin{tabular}{|l<{}|c|>{$}c<{$}|c| |c| |c|}
                                 \hline
    \textbf{Model} & \textbf{Category} & \textbf{Precision} & \textbf{Recall} & \textbf{F1-score} & \textbf{MCC}\\\hline
   BERT & neutral  & 0.78  & 0.83 & 0.80 & 0.47 \\\hline
   BERT & positive  & 0.72  & 0.67 & 0.69 & 0.56 \\\hline
   BERT & negative  & 0.45  & 0.34 & 0.39 & 0.35 \\\hline
   XLNet & neutral  & 0.80  & 0.82 & 0.81 & 0.51\\\hline
   XLNet & positive  & 0.72  & 0.73 & 0.72 & 0.59 \\\hline
   XLNet & negative  & 0.48  & 0.38 & 0.42 & 0.39 \\\hline
   RoBERTa & neutral  & 0.81  & 0.82 & 0.81 & 0.53 \\\hline
   RoBERTa & positive & 0.73  & 0.74 & 0.73 & 0.6 \\\hline  
   RoBERTa & negative  & 0.52  & 0.43 & 0.47 & 0.43 \\\hline  
    CT-BERT & neutral  & 0.81  & 0.83 & 0.82 & 0.49 \\\hline
    CT-BERT & positive  & 0.73  & 0.71 & 0.72 & 0.56 \\\hline
    CT-BERT & negative  & 0.55  & 0.47 & 0.51 & 0.46 \\\hline
 BERTweet & neutral  & 0.81  & 0.84 & 0.82 & 0.54 \\\hline
   BERTweet & positive  & 0.75  & 0.73 & 0.74 & 0.62 \\\hline
   BERTweet & negative  & 0.54  & 0.41 & 0.47 & 0.43 \\\hline
  \end{tabular}
\end{table}

Both Table 8 and Fig. 3 indicate a decrease in the performance scores (after text oversampling) of the neutral class (majority class) as compared to the results of the original dataset in Table 5 and Fig. 2. The positive class accuracies remain more or less the same with a slight increase noted for some models. However, the negative class scores have distinctly decreased for all models which can be attributed to text oversampling using the limited number of tweets available in the negative class.

\subsection{Discussion}
In our work, we investigate the utility of domain-specific pre-trained transformer models and text oversampling for the sentiment analysis of Covid-19 vaccine related tweets from an imbalanced small sample dataset. As observed from the performance scores in Tables 4-8, the domain-specific pre-trained transformer models CT-BERT and BERTweet significantly outperform the pre-trained transformer models - RoBERTa, BERT and XLNet. An instance of a Covid-19 tweet that is classified correctly by the domain-specific CT-BERT and BERTweet, but incorrectly by all other transformer models, is shown in Table 9, along with another instance of a tweet which is mis-classified by all the models including CT-BERT and BERTweet. The latter tweet is a mixture of positive and negative news, though the human annotation labelled it as negative. Both CT-BERT and BERTweet labeled the tweet as positive due to the phrase “raised no safety concerns”. 

\begin{table}[!ht]
\centering
\caption{Examples of tweet classification by the pre-trained transformer models}
\label{tabl.1}
  \begin{tabular} {|>{\centering\arraybackslash}m{2.75cm}|>{\centering\arraybackslash}m{1cm}|>{\centering\arraybackslash}m{1cm}|>{\centering\arraybackslash}m{1cm}|>{\centering\arraybackslash}m{1.5cm}|>{\centering\arraybackslash}m{1cm}|>{\centering\arraybackslash}m{1.75cm}|}
                                 \hline
    \textbf{Tweet}   & \textbf{ground truth}  &\textbf{BERT} & \textbf{XLNet}  &\textbf{RoBERTa}  &\textbf{CT-BERT}  &\textbf{BERTweet} \\\hline
i will not be taking amp j or any other vaccine. it's clear now that governments have no idea of the safety profile short medium long term of these experimental vaccines   & negative & neutral & neutral & neutral & negative & negative \\\hline
there were six deaths during the late stage trials but the fda says this raised no safety concerns & negative & neutral & neutral & neutral & positive & positive \\\hline
  \end{tabular}
\end{table}

The following observations were made from the performance scores of the five transformer models before and after text oversampling by LMOTE (augmentation of the minority classes only).

1. The domain-specific Covid-Twitter-BERT (CT-BERT) model performed significantly better than the pre-trained models- RoBERTa, XLNet and BERT for the original dataset (Table 4) since it was pre-trained on a Twitter dataset consisting of only Covid-19 tweets. The CT-BERT results also outperformed the other domain-specific model BERTweet for the original non-augmented dataset.

2. BERTweet, which has BERT as the base model, and is trained on English Tweets, gives consistently better results, and outperformed CT-BERT for the augmented dataset. The results of CT-BERT and BERTweet were overall better than that of RoBERTa, XLNet and BERT, which is expected since they both are trained on domain-specific information (tweets related to the Covid-19 pandemic).

3. Both CT-BERT and BERTweet performed well for the minority class containing negative sentiments even without text augmentation (Table 5), indicating that pre-training with domain-specific information helps to mitigate the effects of an imbalanced class distribution.

3. RoBERTa model was proved to be a better fit for the task at hand as compared to BERT and XLNet as observed from the accuracy, F1-score, MCC scores in Tables 4 and 6.

4. Training the models using synthetically generated textual data yielded worse results for the neutral class, and a marginal increase for the positive class, while the scores of the negative class significantly decreased as observed from Tables 5 and 8. 

5. It was concluded that LMOTE works poorly in multi-class settings, often degrading performance for the majority classes. LMOTE was used to augment and balance the dataset, in our case, so that training can be performed equally for all classes.

6. Thus text oversampling is not an advisable choice in the case of an imbalanced small sample multi-class dataset since it can downgrade the precision and/or recall rate for the majority class, as observed from the drop in the performance scores of the neutral class in Table 8. Though oversampling does improve results in data mining, the results of oversampling of Covid-19 tweets using LMOTE and SMOTE (Tables 6 and 7) were not encouraging, since the synthetically generated text for the low population negative class was not of high quality and degraded the performance of the neutral class (majority class).

\section{Conclusion}

In this paper, we explore the effectiveness of domain-specific pre-trained transformer models and text oversampling for learning from small sample datasets with an imbalanced class distribution. We consider the specific task of sentiment analysis of Covid-19 vaccine-related tweets. The majority class is the neutral sentiment, while the positive and negative sentiments form the minority classes. The performance scores of the negative sentiment class are the lowest which occurred due to the small number of training samples in this class. In this scenario, the domain-specific pre-trained transformer models CT-BERT and BERTweet outperform RoBERTa, BERT and XLNet transformer models that are state-of-the-art pre-trained transformer models popularly used for text classification tasks. Thus we conclude that domain-specific transformer models are able to mitigate the class imbalance to a certain extent. Text oversampling of the minority class Covid-19 tweets was found to deteriorate the overall performance, with  BERTweet performing better than the other models on the augmented dataset. Hence synthetic tweet generation via text oversampling for the minority classes is not advisable for imbalanced small sample text datasets. We propose to adapt domain-specific transformer models for the classification of Covid-19 related documents in digital repositories in our future works. Since both CT-BERT and BERTweet models are based on the BERT transformer model, we would like to explore, in future, the pre-training of more recent transformer versions such as XLNet using Covid-19 related tweets.

\section{References}

[1] URL https://www.kaggle.com/gpreda/all-covid19-vaccines-tweets. Last
accessed on 11th Feb 2022.

[2] Adeyemi I., Esan A.: Covid-19-Related Health Information Needs and Seeking
Behavior among Lagos State Inhabitants of Nigeria. In: International Journal of
Information Science and Management (IJISM), vol. 20(1), 2022.

[3] Adoma A., Henry N., Chen W.: Comparative analyses of bert, roberta, distil-bert, and xlnet for text-based emotion recognition. In: 2020 17th International
Computer Conference on Wavelet Active Media Technology and Information Processing (ICCWAMTIP), p. 117–121. IEEE, 2020.

[4] Al-Hashedi A., Al-Fuhaidi B., Mohsen A.M., Ali Y., Gamal Al-Kaf H.A., Al-Sorori W., Maqtary N.: Ensemble classifiers for Arabic sentiment analysis of
social network (Twitter data) towards COVID-19-related conspiracy theories.
In: Applied Computational Intelligence and Soft Computing, vol. 2022, 2022.

[5] Alenezi M.N., Alqenaei Z.M.: Machine learning in detecting COVID-19 misinformation on twitter. In: Future Internet, vol. 13(10), p. 244, 2021.

[6] Araci D.: Finbert: Financial sentiment analysis with pre-trained language models, 2019. ArXiv preprint arXiv:1908.10063.

[7] Bahdanau D., Cho K., Bengio Y.: Neural machine translation by jointly learning
to align and translate. In: 3rd International Conference on Learning Representations, ICLR. 2015.

[8] Bansal A., Susan S., Choudhry A., Sharma A.: Covid-19 Vaccine Sentiment
Analysis During Second Wave in India by Transfer Learning Using XLNet. In:
International Conference on Pattern Recognition and Artificial Intelligence, pp.
443–454. Springer, 2022.

[9] Beltagy I., Lo K., Cohan A.: SciBERT: A Pretrained Language Model for Scientific Text. In: Proceedings of the 2019 Conference on Empirical Methods in
Natural Language Processing and the 9th International Joint Conference on Natural Language Processing (EMNLP-IJCNLP), p. 3615–3620. 2019.

[10] Chawla N.V., Bowyer K.W., Hall L.O., Kegelmeyer W.P.: SMOTE: synthetic minority over-sampling technique. In: Journal of artificial intelligence research,
vol. 16, pp. 321–357, 2002.

[11] Dastgheib M., Koleini S., Rasti F.: The application of deep learning in persian
documents sentiment analysis. In: International Journal of Information Science
and Management (IJISM), vol. 18(1), p. 1–15, 2020.

[12] Devlin J., Chang M., Lee K., Toutanova K.: BERT: Pre-training of Deep Bidirectional Transformers for Language Understanding. In: Proceedings of the 2019
Conference of the North American Chapter of the Association for Computational
Linguistics: Human Language Technologies, vol. 1, p. 4171–4186. 2019.

[13] Goel R., Susan S., Vashisht S., Dhanda A.: Emotion-Aware Transformer Encoder
for Empathetic Dialogue Generation. In: 2021 9th International Conference on
Affective Computing and Intelligent Interaction Workshops and Demos (ACIIW),
p. 1–6. IEEE, 2021.

[14] Huang K., Altosaar J., Ranganath R.: Clinicalbert: Modeling clinical notes and
predicting hospital readmission, 2019. ArXiv preprint arXiv:1904.05342.

[15] Hutto C., Gilbert E.: Vader: A parsimonious rule-based model for sentiment
analysis of social media text. In: Proceedings of the international AAAI conference on web and social media, vol. 8(1), p. 216–225, 2014.

[16] Kou G., Yang P., Peng Y., Xiao F., Chen Y., Alsaadi F.: Evaluation of feature
selection methods for text classification with small datasets using multiple criteria
decision-making methods. In: Applied Soft Computing, vol. 86, p. 105836, 2020.

[17] Lan Z., Chen M., Goodman S., Gimpel K., Sharma P., Soricut R.: Albert: A
lite bert for self-supervised learning of language representations, 2019. ArXiv
preprint arXiv:1909.11942.

[18] Lee J., Hsiang J.: Patentbert: Patent classification with fine-tuning a pre-trained
bert model, 2019. ArXiv preprint arXiv:1906.02124.

[19] Lee J., Yoon W., Kim S., Kim D., Kim S., So C., Kang J.: BioBERT: a pre-trained biomedical language representation model for biomedical text mining. In:
Bioinformatics, vol. 36(4), p. 1234–1240, 2020.

[20] Leekha M., Goswami M., Jain M.: A multi-task approach to open domain suggestion mining using language model for text over-sampling. In: European Conference on Information Retrieval, p. 223–229. Springer, Cham, 2020.

[21] Liew T., Lee C.: Examining the Utility of Social Media in COVID-19 Vaccination:
Unsupervised Learning of 672,133 Twitter Posts. In: JMIR public health and
surveillance, vol. 7(11), p. 29789, 2021.

[22] Liu J., Lu S., Lu C.: Exploring and Monitoring the Reasons for Hesitation with
COVID-19 Vaccine Based on Social-Platform Text and Classification Algorithms.
In: Healthcare, vol. 9, p. 1353. MDPI), 2021.

[23] Liu S., Liu J.: Public attitudes toward COVID-19 vaccines on English-language
Twitter: A sentiment analysis. In: Vaccine, vol. 39(39), p. 5499–5505, 2021.

[24] Liu Y., Ott M., Goyal N., Du J., Joshi M., Chen D., Stoyanov V.: 2019. Roberta:
A robustly optimized bert pretraining approach. arXiv preprint arXiv:1907.11692.

[25] Lu J., Plataniotis K., Venetsanopoulos A.: Regularization studies of linear discriminant analysis in small sample size scenarios with application to face recognition. In: Pattern recognition letters, vol. 26(2), p. 181–191, 2005.

[26] Mallick R., Susan S., Agrawal V., Garg R., Rawal P.: Context-and sequence-aware convolutional recurrent encoder for neural machine translation. In: Proceedings of the 36th Annual ACM Symposium on Applied Computing, p. 853–856.
2021.

[27] Manguri K., Ramadhan R., Amin P.: Twitter sentiment analysis on worldwide
COVID-19 outbreaks. In: Kurdistan Journal of Applied Research, p. 54–65, 2020.

[28] Marcec R., Likic R.: Using twitter for sentiment analysis towards AstraZeneca/Oxford, Pfizer/BioNTech and Moderna COVID-19 vaccines. In: Post-graduate Medical Journal, 2021.

[29] Martin L., Muller B., Su´arez P., Dupont Y., Romary L., De La Clergerie V.,
Sagot B.: CamemBERT: a Tasty French Language Model. In: Proceedings of
the 58th Annual Meeting of the Association for Computational Linguistics, p.
7203–7219. 2020.

[30] M¨uller M., Salath´e M., Kummervold P.: Covid-twitter-bert: A natural language
processing model to analyse covid-19 content on twitter, 2020. ArXiv preprint
arXiv:2005.07503.

[31] Mohsen A., Ali Y., Al-Sorori W., Maqtary N.A., Al-Fuhaidi B., Altabeeb A.M.:
A performance comparison of machine learning classifiers for Covid-19 Arabic
Quarantine tweets sentiment analysis. In: 2021 1st International Conference on
Emerging Smart Technologies and Applications (eSmarTA), pp. 1–8. IEEE, 2021.

[32] Naseem U., Razzak I., Khushi M., Eklund P., Kim J.: COVIDSenti: A large-scale benchmark Twitter data set for COVID-19 sentiment analysis. In: IEEE
Transactions on Computational Social Systems, vol. 8(4), p. 1003–1015, 2021.

[33] Naseem U., Razzak I., Musial K., Imran M.: Transformer based deep intelligent contextual embedding for twitter sentiment analysis. In: Future Generation
Computer Systems, vol. 113, p. 58–69, 2020.

[34] Nguyen D., Vu T., Nguyen A.: BERTweet: A pre-trained language model for
English Tweets. In: Proceedings of the 2020 Conference on Empirical Methods
in Natural Language Processing: System Demonstrations, p. 9–14. 2020.

[35] Nowak S., Chen C., Parker A., Gidengil C., Matthews L.: Comparing covariation
among vaccine hesitancy and broader beliefs within Twitter and survey data. In:
PloS one, vol. 15(10), p. 0239826, 2020.

[36] Olaleye T., Abayomi-Alli A., Adesemowo K., Arogundade O.T., Misra S., Kose
U.: SCLAVOEM: hyper parameter optimization approach to predictive modelling
of COVID-19 infodemic tweets using smote and classifier vote ensemble. In: Soft
Computing, pp. 1–20, 2022.

[37] Pires T., Schlinger E., Garrette D.: How Multilingual is Multilingual BERT?
In: Proceedings of the 57th Annual Meeting of the Association for Computational
Linguistics, p. 4996–5001. 2019.

[38] Saini M., Susan S.: Data augmentation of minority class with transfer learning for
classification of imbalanced breast cancer dataset using inception-V3. In: Iberian
Conference on Pattern Recognition and Image Analysis, p. 409–420. Springer,
Cham, 2019.

[39] Sattar N., Arifuzzaman S.: COVID-19 Vaccination awareness and aftermath:
Public sentiment analysis on Twitter data and vaccinated population prediction
in the USA. In: Applied Sciences, vol. 11(13), p. 6128, 2021.

[40] Scheible R., Thomczyk F., Tippmann P., Jaravine V., Boeker M.: Gottbert: a
pure german language model, 2020. ArXiv preprint arXiv:2012.02110.

[41] Somasundaran S.: Two-level transformer and auxiliary coherence modeling for
improved text segmentation. In: Proceedings of the AAAI Conference on Artificial Intelligence, vol. 34(05), p. 7797–7804, 2020.

[42] Susan S., Kumar A.: The balancing trick: Optimized sampling of imbalanced
datasets—A brief survey of the recent State of the Art. In: Engineering Reports,
vol. 3(4), p. 12298, 2021.

[43] Vashishtha S., Susan S.: Inferring sentiments from supervised classification of
text and speech cues using fuzzy rules. In: Procedia Computer Science, vol. 167,
p. 1370–1379, 2020.

[44] Vaswani A., Shazeer N., Parmar N., Uszkoreit J., Jones L., Gomez A., Polosukhin
I.: Attention is all you need. In: Advances in neural information processing
systems, vol. 30, 2017.

[45] Wang T., Lu K., Chow K., Zhu Q.: COVID-19 Sensing: Negative sentiment
analysis on social media in China via Bert Model. In: Ieee Access, vol. 8, p.
138162–138169, 2020.

[46] Yang Z., Dai Z., Yang Y., Carbonell J., Salakhutdinov R., Le Q.: Xlnet: Generalized autoregressive pretraining for language understanding, 2019. Advances in
neural information processing systems, 32.

\end{document}